\pdfoutput=1

\documentclass[11pt]{article}

\usepackage{acl}
\usepackage{graphicx}
\usepackage{booktabs}
\usepackage{float}
\usepackage{bm}
\usepackage{amsmath}
\usepackage{multirow} 
\usepackage{times}
\usepackage{latexsym}

\usepackage{color}
\usepackage{xcolor}

\usepackage{algorithm}
\usepackage{algpseudocode}

\usepackage{CJKutf8}

\DeclareMathOperator*{\argmax}{arg\,max}

\usepackage[T1]{fontenc}

\usepackage[utf8]{inputenc}

\usepackage{microtype}

\title{MGIMN: Multi-Grained Interactive Matching Network for \\ Few-shot Text Classification}

\author{
Jianhai Zhang$^{1}$, 
Mieradilijiang Maimaiti$^{1}$,
Gao Xing$^{1}$,
Yuanhang Zheng$^{2,3,4}$,
and Ji Zhang$^{1, *}$
\\
$^1$Alibaba DAMO Academy \\
$^2$Department of Computer Science and Technology, Tsinghua University, Beijing, China \\
$^3$Institute for Artificial Intelligence, Tsinghua University, Beijing, China \\
$^4$Beijing National Research Center for Information Science and Technology \\
\tt{\{tanfan.zjh,mieradilijiang.mea,gaoxing.gx,zj122146\}@alibaba-inc.com}\\
\tt{zheng-yh19@mails.tsinghua.edu.cn}
}

\begin{document}
\begin{CJK}{UTF8}{gbsn}

\maketitle
\begin{abstract}
Text classification struggles to generalize to unseen classes with very few labeled text instances per class.
In such a few-shot learning (FSL) setting, metric-based meta-learning approaches have shown promising results. 
Previous studies mainly aim to derive a prototype representation for each class.
However, they neglect that it is challenging-yet-unnecessary to construct a compact representation which expresses the entire meaning for each class.
They also ignore the importance to capture the inter-dependency between query and the support set for few-shot text classification. 
To deal with these issues, we propose a meta-learning based method \textbf{MGIMN} which performs instance-wise comparison followed by aggregation to generate class-wise matching vectors instead of prototype learning.
The key of instance-wise comparison is the interactive matching within the class-specific context and episode-specific context. 
Extensive experiments demonstrate that the proposed method significantly outperforms the existing SOTA approaches, under both the standard FSL and generalized FSL settings.
\end{abstract}

{
\let\thefootnote\relax\footnotetext{
$^{*}$ Corresponding author: Ji Zhang}
}

\section{Introduction}

 Few-shot text classification has attracted considerable attention because of significant academic research value and practical application value \citep{gao2019hybrid,yin2020meta,brown2020language,bragg2021flex,liu2021pre}. Many efforts are devoted towards different goals like generalization to new classes \citep{gao2019hybrid,ye2019multi,nguyen2020dynamic,  bragg2021flex}, adaptation to new domains and tasks \citep{bansal2020self,brown2020language, schick2020s, gao2020making, bragg2021flex}. Low-cost generalization to 
new classes is critical to deal with the growing long-tailed categories, which is common for intention classification \citep{geng2019induction, krone2020learning} and relation classification \citep{han2018fewrel, gao2019hybrid}.

\begin{figure}[!t]
\centering

\includegraphics[width=0.48\textwidth,height=5.1cm]{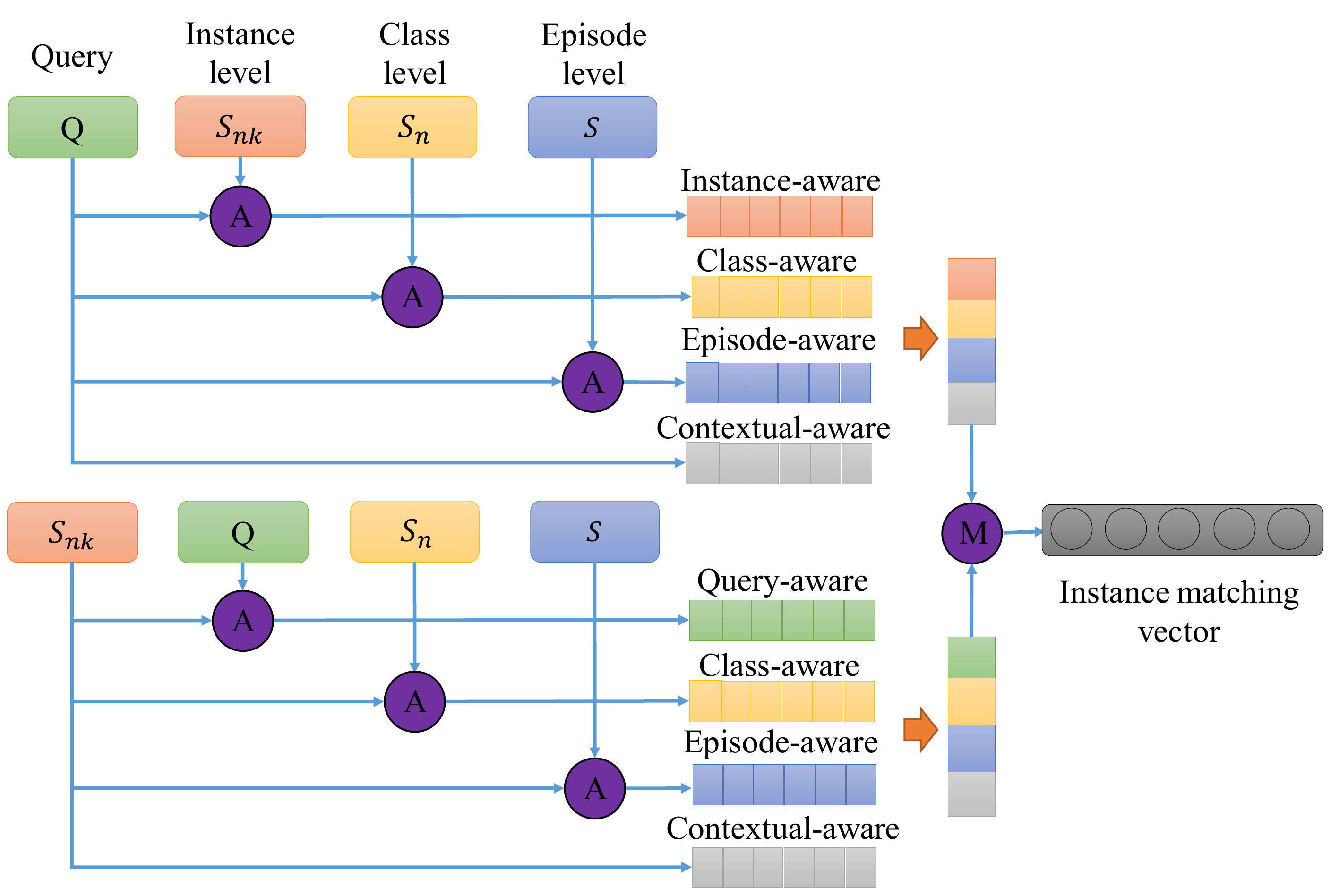}

\caption{The example of multi-grained few-shot text classification. The ``$S_{nk},S_n$'' and ``$S$'' represent the instance level, class level and episode level interaction respectively. While ``A'' in the purple circle denotes alignment between query and instances. The ``'M'' stands for matching operation.} 
\label{fig_sample_mgimn}
\end{figure}

To prevent over-fitting to few-shot new classes and avoid retraining the model when the class space changes, metric-based meta learning has become the major framework with significant results \citep{yin2020meta,bansal2020self}. The core idea is that episode sampling is employed in meta-training phase to learn the relationship between query and candidate classes \citep{bragg2021flex}. A key challenge is inducing class-wise representations from support sets because nature language expressions are diverse \citep{gao2019hybrid} and  highly informative lexical features are unshared across episodes \citep{bao2019few}.

Under metric-based meta learning framework, many valuable backbone networks have emerged. \citet{snell2017prototypical} presented prototypical network that computes prototype of each class using a simple mean pooling method. \citet{gao2019hybrid} proposed hybrid attention mechanism to ease the negative effects of noisy support examples. \citet{geng2019induction} proposed an induction network to induce better prototype representations. \citet{ye2019multi} obtained each prototype vector by aggregating local matching and instance matching information. \citet{bao2019few} proposed a novel method to compute prototypes based on both lexical features and word occurrence patterns. All these previous works first obtain class-wise representations and then perform class-wise comparisons. However, it is challenging-yet-unnecessary to construct a compact representation which expresses the entire meaning for each class \citep{parikh2016decomposable}. In text matching research, compare-aggregate methods which perform token-level comparisons followed by sentence-level aggregation has already been successful \citep{parikh2016decomposable,tay2017compare, yang2019simple}. 
Besides backbone networks,there are still some work that can be further combined. \citet{luo2021don} utilized class-label information for extracting more discriminative prototype representation. \citet{bansal2020self}  generated a large-scale meta tasks from unlabeled text in a self-supervised manner.

In this paper we propose \textbf{M}ulti-\textbf{g}rained \textbf{I}nteractive \textbf{M}atching \textbf{N}etwork, a backbone network for few-shot text classification. The core difference between us with previous efforts is that our work performs instance-level comparisons followed by class-wise aggregation. Specifically, first, all text sequences including query and all support instances are encoded to contextual representations. Second, as depicted in Figure \ref{fig_sample_mgimn}, we design a novel multi-grained interactive matching mechanism to perform instance-wise comparisons which capture the inter-dependency between query and all support instances. Third, class-wise aggregate layer obtains class-wise matching vector between query and each class. Finally, a prediction layer predicts final results.
In contrast to standard FSL setting, generalized FSL setting is a more challenging-yet-realistic setting where seen classes and new classes are co-existent \citep{nguyen2020dynamic,li2020boosting}. In such a  setting, we analyze the relationship between inference speed and the number of classes, and verify the necessity of retrieval, which is ignored by previous studies.

Our contributions are listed as follows:
\begin{itemize}
    \item We propose \textbf{MGIMN} which is more concerned with matching features than semantic features through multi-grained interactive matching.
    \item We verify the necessity of retrieval for realistic applications of few-shot text classification when the number of classes grows.
    \item We conduct extensive experiments and achieve SOTA results under both the standard FSL and generalized FSL settings. 
\end{itemize}


\begin{figure*}[!t]
\centering

\includegraphics[width=0.75\textwidth,height=9.0cm]{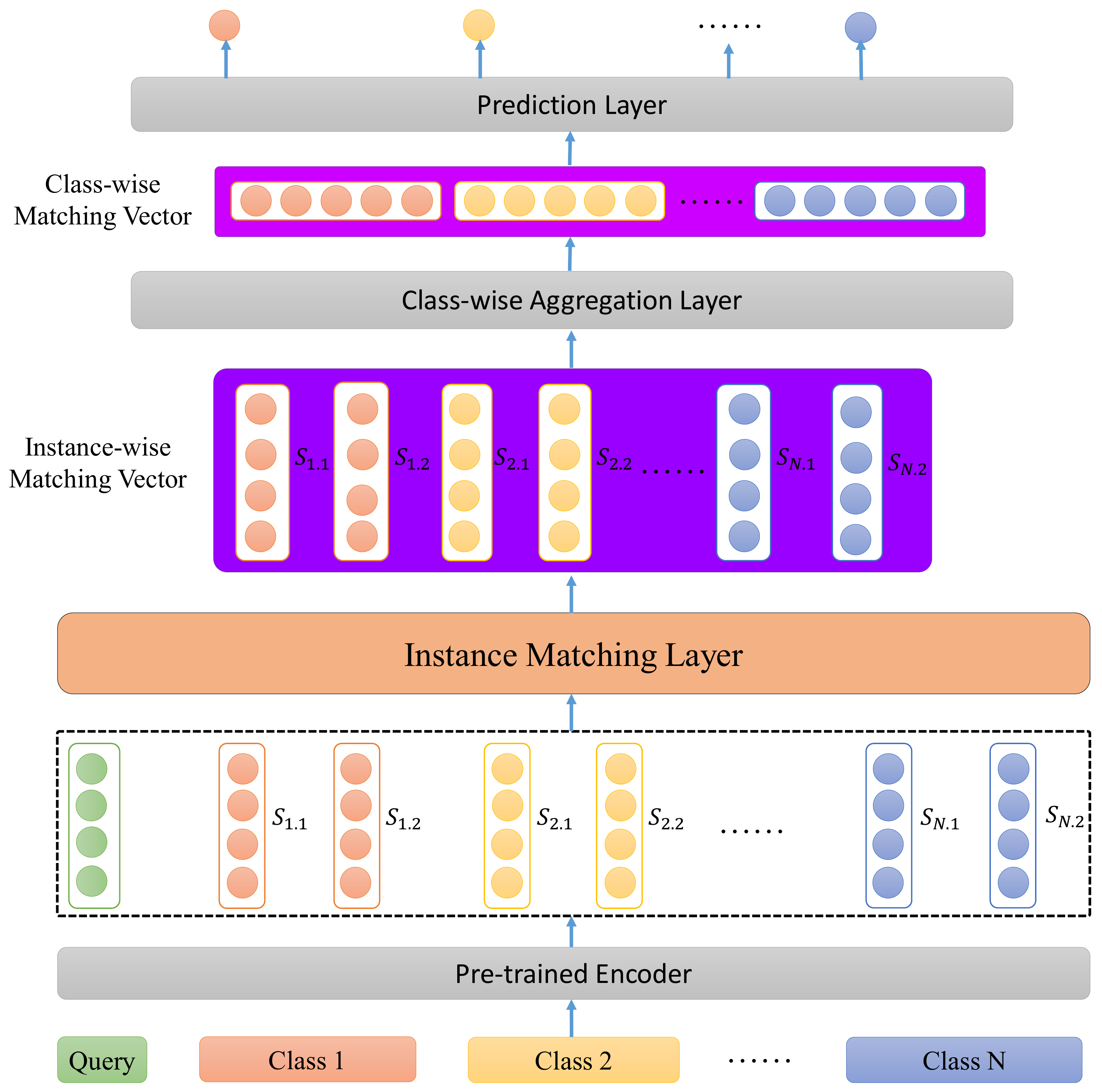}
\caption{The main architecture of multi-grained few-shot text classification model. The details of ``Instance Matching Layer'' totally same as depicted as Figure \ref{fig_sample_mgimn}.} 
\label{fig_framework}
\end{figure*}

\section{Background}\label{task_def}

\subsection{Few-Shot Learning}

Few-shot learning focuses on construct a classifier $G(S,q)\rightarrow y$ which maps the query $q$ to the label $y$ given the support set $S$. Few-shot learning is significantly different from traditional machine learning. In traditional machine learning tasks, we directly sample batches of training data during training. Unlike traditional machine learning models, few-shot learning models usually adopt episode training strategy which means we build meta tasks using sampled training data in each training step.

Specifically, during each training step, $N$ different classes are randomly chosen. After the classes are chosen, we sample $R$ samples as query set $Q$ and $K$ samples for each class as support set $S$ where $Q\bigcap{S}\in\emptyset$. For training, given the support set $S=\{S_n^{k};i=1,..,N,j=1,..,K\}$ , and query set $Q=\{(q_i,y_i); i=1,2,..,R,  y_i\in1,..,N\}$ which has ${R}$ samples in each training step,  the training objective is to minimize:
\begin{equation}
  J=-\frac{1}{R} \sum_{(q,y)\in Q}log(P(y|(S,q))).
\end{equation}

For evaluation, as we described in introduction section, there are two settings. In standard FSL setting, we do N-way K-shot sampling on the classes for validation and test (which are unseen during training) to construct episodes for validation and test.  
In generalized FSL setting, we reformulate the FSL task as ${C}$-way ${K}$-shot classification where C is the count of all classes for training, validation and test, and usually is far greater than N. In this setting, we construct episodes by sampling on all classes for test.

\subsection{Matching Network}

Matching Network \citep{vinyals2016matching} is a typical few-shot learning approach, which leverages the cosine similarity to perform few-shot classification. Specifically, for the query instance $q$ and each support instance $S_n^k \in S$, the cosine similarity between $q$ and $S_n^k$ is computed as follow:
\begin{equation}
    sim(q,S_n^k)=\frac{q \cdot S_n^k}{||q||\ ||S_n^k||}.
\end{equation}

Then, we compute the probability distribution of the label $y$ of the query $q$ using attention:
\begin{equation}
    P(y|S,q)=\frac{\sum_{k=1}^{K}\exp(sim(q,S_y^k))}{\sum_{n=1}^{N}\sum_{k=1}^{K}\exp(sim(q,S_n^k))}.
\end{equation}

Finally, for any query instance $q$, we regard the class with the maximum probability as its label $y$:
\begin{equation}
    y=\mathop{\argmax}_{n}P(y=n|S,q).
\end{equation}

\subsection{Text Classification}

As a basic task in NLP, text classification has attracted much attention. In previous works, different model architectures, including RNN \citep{zhou-etal-2016-text} and CNN \citep{DBLP:conf/emnlp/Kim14} are used for text classification. After the appearance of pre-trained language models like BERT \citep{bert}, they have become the mainstream method for text classification. In such methods, the input sentence is encoded into its representation using the Transformer \citep{VaswaniEtAl17-Transformer} architecture through adding $[CLS]$ token before the original input sentence $\mathbf{x}$ and then computing the output representation of $[CLS]$ using the model. 
\begin{equation}
    \mathbf{h}_{\text{CLS}}=Transformer(\text{[CLS]},\mathbf{x};\theta),
\end{equation}

\noindent where $\theta$ represents the model's parameters.

Then, according to the representation, the probability distribution of $y$ can be computed as follow:
\begin{equation}
    P(y|\mathbf{x};\theta)=softmax(\mathbf{W}_{\text{softmax}}\mathbf{h}_{\text{CLS}}),
\end{equation}

\noindent where $\mathbf{W}_{\text{softmax}}$ is the parameters of the softmax layer.

\section{Method}
As illustrated in the Figure \ref{fig_framework}, MGIMN consists of four modules:  Encoder Layer, Instance Matching Layer, Class-wise Aggregation Layer and Prediction Layer.


\subsection{Encoder Layer}
We employ transformer encoder from pre-trained BERT as encoder layer. Similar to the original work, we add a special token [CLS] before original text. Then the encoder layer takes a token sequence as input and outputs token-wise sequence representation. Instead of using the vector of [CLS] token as sentence-wise representation, we adopt final hidden states of the rest tokens for further fine-grained instance-wise matching.

We denote $x=\{w_1,w_2,...,w_l\}$ as a token sequence. Encoder Layer outputs the token-wise representation $h=\{h_1,h_2,...,h_l\}$, where $l$ denotes length of the token sequence. Query and each support instance are encoded individually. We denote ${q}$ as encoded result of query and ${s_{nk}}$ as encoded result of the ${k_{th}}$ support instance of ${n_{th}}$ class.

\subsection{Instance Matching Layer} 
This is the core of our model. Instance-wise matching vectors are obtained by comparing query with each support instance.  

\subsubsection{Bidirectional Alignment}
Following previous works\citep{parikh2016decomposable, yang2019simple} in text matching , we use bidirectional alignment to capture inter-dependency between two text sequences. 

\begin{equation}
\widehat{a}, \widehat{b} =BiAlign(a,b)
\end{equation}

\noindent where a and b denote the token-wise sequence representations and ${BiAlign}$ denotes the bidirectional alignment function defined as follows:

\begin{equation}
e_{ij}=\mathbf{F}(a_i)^{\text{T}}\mathbf{F}(b_j) 
\end{equation}

\begin{equation}
\widehat{a_i}=\sum \limits^{l_b}_{j=1}\frac{exp(e_{ij})}{\sum ^{l_b}_{k=1} exp(e_{ik})}b_j 
\label{interaction_formula}
\end{equation}
\begin{equation}
\widehat{b_j}=\sum \limits^{l_a}_{i=1}\frac{exp(e_{ij})}{\sum ^{l_a}_{k=1} exp(e_{kj})}a_i 
\end{equation}

\noindent where $a_i$ and $b_j$ denotes representation of $i_{th}$ and $j_{th}$ token of $a$ and $b$, respectively, $\widehat{a_i}$ and $\widehat{b_j}$ denote the aligned representations, and $F$ is a single-layer feed forward network.

\subsubsection{Multi-grained Interactive Matching} 
In few-shot text classification, judging whether query and each support instance belong to the same category cannot be separated from class-specific context and episode-specific context. There are three components including alignment, fusion and comparison. 

For alignment, besides local alignment between query and each support instance, we also consider their alignments with global context information. We denote ${S_n} = concat(\{s_{nj}\}_{j=1}^K)$ as class-specific context and $S = concat(\{S_i\}_{i=1}^N)$ as episode-specific context. The multi-grained alignments for query and each support instance are performed as follows:
 \begin{equation}
 \begin{aligned}
        q_{nk}^{'},s_{nk}^{'}=BiAlign(q,{s_{nk}}) \\
        q_n^{''},\_=BiAlign(q,S_n) \\
        s_{nk}^{''},\_=BiAlign({s_{nk}},S_n)\\
        q^{'''},\_=BiAlign(q,S) \\
        s_{nk}^{'''},\_=BiAlign({s_{nk}},S)
\end{aligned}
\end{equation}
where $q_{nk}^{'}$, $q_n^{''}$ and $q^{'''}$ denote instance-aware, class-aware and episode-aware query representations respectively, $s_{nk}^{'}$, $s_{nk}^{''}$ and $s_{nk}^{'''}$ denote query-aware, class-aware and episode-aware support instance representations respectively.

For fusion, we fuse original representation and three kinds of aligned representations together as follows:
\begin{equation}
\begin{aligned}
\bm{q_{nk}^{'}}=H_1(q;q_{nk}^{'};\left| q-q_{nk}^{'} \right|;q\odot{q_{nk}^{'}})\\
\bm{q_{nk}^{''}}=H_2(q;q_n^{''};\left| q-q_n^{''} \right|;q\odot{q_n^{''}})\\
\bm{q_{nk}^{'''}}=H_3(q;q^{'''};\left| q-q^{'''} \right|;q\odot{q^{'''}})\\
\bm{s_{nk}^{'}}=H_1({s_{nk}};s_{nk}^{'};\left| {s_{nk}}-s_{nk}^{'} \right|;{s_{nk}}\odot{s_{nk}^{'}})\\
\bm{s_{nk}^{''}}=H_2({s_{nk}};s_{nk}^{''};\left| {s_{nk}}-s_{nk}^{''} \right|;{s_{nk}}\odot{s_{nk}^{''}})\\
\bm{s_{nk}^{'''}}=H_3({s_{nk}};s_{nk}^{'''};\left| {s_{nk}}-s_{nk}^{'''} \right|;{s_{nk}}\odot{s_{nk}^{'''}})\\
\label{fusion}
\end{aligned}
\end{equation}
\begin{equation}
\begin{aligned}
   \bm{q_{nk}}=H(\bm{q_{nk}^{'}};\bm{q_{nk}^{''}};\bm{q_{nk}^{'''}}) \\
   \bm{s_{nk}}=H(\bm{s_{nk}^{'}};\bm{s_{nk}^{''}};\bm{s_{nk}^{'''}})
\end{aligned}
\end{equation}
where $H_1,H_2,H_3,H$ are feed forward networks with single-layer and initialize with independent parameters, $\odot$ denotes the element-wise multiplication operation, and $;$ denotes concatenation operation.

For comparison, the instance-wise matching vector is computed as follows:
\begin{equation}
\begin{aligned}
    \vec{q_{nk}}= [max(\{\bm{q_{nk}}\});avg(\{\bm{q_{nk}}\})]\\
    \vec{s_{nk}}= [max(\{\bm{s_{nk}}\});avg(\{\bm{s_{nk}}\})]\\
    \vec{m_{nk}}= G(\vec{q_{nk}};\vec{s_{nk}};\left|\vec{q_{nk}}-\vec{s_{nk}}\right|;\vec{q_{nk}}\odot\vec{s_{nk}})
\end{aligned}
\end{equation}
where  $\bm{q_{nk}}$ is a $l_q\times{D}$ matrix, $\bm{s_{nk}}$ is a $l_s\times{D}$ matrix, $\vec{q_{nk}}$ and $\vec{s_{nk}}$ are vectors with shape ($1\times{2D}$), $G$ is single-layer feed forward networks, $l_q$ and $l_s$ are sequence length of query and support instance respectively, and $D$ is hidden size.

\subsection{Class Aggregation Layer}
This layer aggregate instance-wise matching vectors into class-wise matching vectors for final prediction. 
\begin{equation}
        \vec{c_n}=[max(\{\vec{m_{nk}}\}_{k=1}^K);avg(\{\vec{m_{nk}}\}_{k=1}^K)]
        \label{class_agg}
\end{equation}
where $\vec{c_n}$ denotes the final matching vector of $n_{th}$ class, ``$;$'' denotes the concatenation of two vectors and $\vec{m_{nk}}$ denotes instance-wise matching vector produced by instance matching layer.

\subsection{Prediction Layer}
Finally, prediction layer, which is a two-layer fully connected network with the output size is $1$, is applied to the matching vector $\vec{c_n}$ and outputs final predicted result.
\begin{equation}
        logit_n= MLP(\vec{c_n}), n=1,..,N
\end{equation}

\section{Experiments}

\begin{table*}[]
\resizebox{\textwidth}{!}{
\begin{tabular}{lcc|ccccccc}
\hline
\multirow{2}{*}{Datasets}  & \multicolumn{2}{c}{Standard FSL Setting} & \multicolumn{7}{c}{Generalized FSL Setting} \\ \cline{2-10} 
 & \#sentences & $C_{tr}$/$C_{val}$/$C_{test}$ & C & K & \#sc & \#uc & \#$D_{tr}$ & \#$D_{val}$ & \#$D_{test}$ \\ \hline
OOS & 22500 & \multicolumn{1}{c|}{50/50/50} & 150 & 5 & 50 & 100 & 7000 & 1250 & 1250 \\
Liu & 25478 & \multicolumn{1}{c|}{18/18/18} & 54 & 5 & 18 & 36 & 8312 & 450 & 450 \\
Amzn & 3057 & \multicolumn{1}{c|}{106/106/106} & 318 & 5 & 106 & 212 & 1043 & 530 & 530 \\
Huffpost & 41000 & \multicolumn{1}{c|}{14/13/14} & 41 & 5 & 14 & 27 & 13860 & 340 & 340 \\
FaqIr & 1233 & \multicolumn{1}{c|}{17/16/17} & 50 & 5 & 17 & 33 & 309 & 381 & 381 \\ \hline
\end{tabular}}
\caption{The detailed dataset statistics. 
In standard FSL setting, we cut all classes into trainset/validset/testset according to the ratio with 1:1:1. In generalized FSL setting, we reformulate task as a C-way K-shot classification in which only subset of classes are seen in training phase.}
\label{dataset_s}
\end{table*}

\subsection{Setup}
\subsubsection{The preparation of dataset}
\label{data_pre}
The proposed method has been evaluated on five diverse corpora: OOS \cite{larson-etal-2019-evaluation}, Liu \cite{liu2019benchmarking}, FaqIr \cite{karan2016faqir}, Amzn \cite{amzn} and Huffpost \cite{huffpost}. Among them, OOS, Liu and FaqIr datasets are all intent classification datasets. Amzn dataset is designed for fine-grained classification of product reviews. Huffpost dataset is constructed to identify the types of news based on headlines and short descriptions. 
The dataset characteristics is listed in Table \ref{dataset_s}. 
For the standard FSL setting, we construct the ``support \& query" set by sampling the unique N  classes and K samples each class, and R samples for each of classes, respectively.
We conduct two groups of experiments using $N=[5,10],K=5$ and $R=5$.  In the evaluation phase, we sample $500$ episodes and report the average accuracy.
In generalized FSL setting(GFSL for short), we train the model with episode sampling of 5-way 5-shot. And then we evaluate the model performance with C-way K-shot.

For all experiments, we divide all datasets for $5$ times using different random seeds, just like the way of cross validation, to remove the impact of dataset division. And we conduct $3$ experiments for each model by using different random seeds for model initialization. The final results are reported by averaging $5\times{3}=15$ runs. 

For fair comparison, all models implemented in this paper adopt BERT-Tiny$^{1}$ as encoder layer which is a 2-layer 128-hidden 2-heads version of BERT. Meanwhile, these models initialized their paramaeters using the PTM
published by Google and  fine-tuned during training procedure. 
Besides, the encoder layer and all parameters of other layers are randomly initialized. 
We fix some hyper-parameters with default values such as the hidden size $128$, we also exploit Adam optimizer in all experiment settings. 
The learning rate is tuned from ${1e^{-5}}$ to ${1e^{-4}}$ on validation dataset. 
Dropout with a rate of 0.1 is applied before each fully-connected
layer.
The feed-forward networks described in section 3 (e.g. $F, H_1, H_2, H_3, H$ and $G$) are all single fully-connected layers. The prediction layer is a two-layer fully-connected layer.

{
\let\thefootnote\relax\footnotetext{
$^{1}$ https://github.com/google-research/bert }
}

\begin{table*}[]
\centering
\begin{tabular}{lccccccccc}
\hline
\multirow{2}{*}{Methods} & \multicolumn{3}{c}{OOS} & \multicolumn{3}{c}{Liu} & \multicolumn{3}{c}{FaqIr} \\ \cline{2-10} 
 & $5$-way & $10$-way & GFSL & $5$-way & $10$-way & GFSL & $5$-way & $10$-way & GFSL \\ \hline
Proto & 92.20 & 87.91 & \multicolumn{1}{c|}{61.94} & 82.46 & 73.23 & \multicolumn{1}{c|}{47.66} & 89.83 & 81.56 & 60.78 \\
Matching & 89.78 & 84.41 & \multicolumn{1}{c|}{58.34} & 78.25 & 67.45 & \multicolumn{1}{c|}{41.95} & 86.74 & 78.77 & 53.85 \\
Induction & 80.44 & 70.92 & \multicolumn{1}{c|}{34.00} & 65.58 & 51.56 & \multicolumn{1}{c|}{24.73} & 71.62 & 56.99 & 20.10 \\
Proto-HATT & 92.84 & 89.11 & \multicolumn{1}{c|}{65.52} & 82.38 & 75.29 & \multicolumn{1}{c|}{51.27} & 85.01 & 76.17 & 62.62 \\
MLMAN & 95.99 & 93.41 & \multicolumn{1}{c|}{74.39} & 87.39 & 79.82 & \multicolumn{1}{c|}{57.24} & 94.77 & 89.49 & 74.42 \\ \hline
\textbf{MGIMN(ours)} & \textbf{96.36} & \textbf{94.00} & \multicolumn{1}{c|}{\textbf{76.23}} & \textbf{87.84} & \textbf{80.60} & \multicolumn{1}{c|}{\textbf{57.66}} & \textbf{95.14} & \textbf{90.69} & \textbf{75.81} \\ \hline
\end{tabular}
\caption{Experiment results of standard FSL ($n$-way $5$-shot) and generalized FSL with intent classification datasets (OOS,Liu and FaqIr datasets), while the $n$ is set $5$ and $10$ respectively.}
\label{main_result_1}
\end{table*}

\begin{table}[]
\centering
\begin{tabular}{lccc}
\hline
\multirow{2}{*}{Methods} & \multicolumn{3}{c}{Amzn} \\ \cline{2-4} 
 & $5$-way & $10$-way & \multicolumn{1}{l}{GFSL} \\ \hline
Proto & 78.40 & 69.02 & 41.03 \\
Matching & 75.73 & 64.17 & 38.34 \\
Induction & 64.02 & 50.12 & 20.09 \\
Proto-HATT & 78.05 & 69.00 & 41.81 \\
MLMAN & 85.64 & 79.39 & 46.71 \\ \hline
\textbf{MGIMN(ours)} & \textbf{85.96} & \textbf{80.07} & \textbf{49.46} \\ \hline
\end{tabular}
\caption{Experiment results of standard FSL and generalized FSL settings on Amzn datasets, while the FSL setting is same with Table \ref{main_result_1}.}
\label{main_result_2}
\end{table}

\subsubsection{Baselines}
It is vital to compare the introduced method with some strong baselines with two evaluation metrics mentioned above. Note that we re-implement all methods with the same pre-trained encoder for fairly comparison.
\begin{itemize}
    \item \textbf{Prototypical Network (Proto)}\cite{snell2017prototypical} is the first designed and applied to image classification and has also been used to deal with the  text classification issue in recent studies. 
    \item \textbf{Matching Network (Matching)}\cite{vinyals2016matching}  computes the similarity both on each query and per support samples, and then averages them as final prediction score.
    \item \textbf{Induction network (Induction)}\cite{geng2019induction}  proposes an induction module to induce the prototype by using dynamic routing.
    \item \textbf{Proto-HATT}\cite{gao2019hybrid}  is introduced to deal with the issue of noisy and diverse by leveraging instance-level attention and feature-level attention.
    \item \textbf{MLMAN}\cite{ye2019multi}, can be regarded as one of the variants of Proto, encodes query and support in an interactive way.
\end{itemize}

\subsection{Main results}

\paragraph{Overall Performance} Our key experiment results are given in Table \ref{main_result_1}, \ref{main_result_2} and \ref{main_result_3}. We report the averaged scores over 15 runs (different seen-unseen class splits and random seeds as introduced in section \ref{data_pre}) for each dataset and model. Our method remarkably better than all baselines on the five diverse corpora, especially in more challenging generalized FSL setting: the improvements on Huffpost and Amzn datasets are 2.83\% and 2.75\% respectively.

\paragraph{Generalized FSL} In most studies of text classification \cite{bao2019few,gao2019hybrid} with few-shot manner, 
N-way K-shot accuracy is the standard evaluation metric. There are two problems: (1) The metric is not challenging, usually $N=5$ or $N=10$, much smaller than $C$. We also can see that high scores are often reported in some work\cite{bao2019few,gao2019hybrid}.
 (2) It is unable to 
 reflect the real application scenarios where we usually face the entire class space (both seen classes and unseen classes). 
Consequently, the more challenging generalized FSL evaluation metric is employed to focus on the problems. As shown in Table \ref{main_result_1}, \ref{main_result_2} and \ref{main_result_3}, the performance of generalized FSL evaluation is worse and more challenging than standard FSL. It is very meaningful in realistic scenario and can contribute to the further research. It is noteworthy that, comparing with Proto, our proposed approach 
makes bigger improvement in the challenging generalized FSL metric (GFSL) than the improvements in standard FSL metric(FSL), e.g. OOS dataset: $14.29\%$ of GFSL vs $6.09\%$ of FSL and FaqIr dataset: $15.03\%$ of GFSL vs $9.13\%$ of FSL. 
Obviously, it can be implied from the experiment results that, the presented approach has higher effectiveness among such challenging 
scenarios.

\begin{table}[]
\centering
\begin{tabular}{lccc}
\hline
\multirow{2}{*}{Methods} & \multicolumn{3}{c}{Huffpost} \\ \cline{2-4} 
 & $5$-way & $10$-way & \multicolumn{1}{l}{GFSL} \\ \hline
Proto & 51.57 & 36.74 & 16.47 \\
Matching & 49.77 & 34.28 & 14.18 \\
Induction & 44.69 & 29.35 & 10.40 \\
Proto-HATT & 51.23 & 36.65 & 16.06 \\
MLMAN & 52.76 & 38.22 & 16.78 \\ \hline
\textbf{MGIMN(ours)} & \textbf{54.98} & \textbf{40.12} & \textbf{19.61} \\ \hline
\end{tabular}
\caption{Experiment results of standard FSL  and generalized FSL settings on Huffpost datasets, while the FSL setting is same with Table \ref{main_result_1}.}
\label{main_result_3}
\end{table}

\paragraph{Huffpost Dataset} Samples of the same class are more diverse and scattered on Huffpost. For instance, "green streets are healthy streets", "the real heroes of Pakistan" and "what next for Kurdistan ?" are from the same class:"WORLD NEWS". In this scenario, a single class-wise prototype is difficult to represent the entire class semantic. Interestingly, our approach improves more significantly than other datasets, $2.22\%$ of 5-way 5-shot standard FSL metric,$1.9\%$ of 10-way 5-shot standard FSL metric and $2.83\%$ of generalized FSL metric. In our approach, richer matching features gained through interacting from low level with multi-grained interaction, are effective on the dataset with diverse expressed samples.

\begin{table*}[]
\centering
\begin{tabular}{lcccccc}
\hline
\multirow{2}{*}{Methods} & \multicolumn{3}{c}{Liu} & \multicolumn{3}{c}{Huffpost} \\ \cline{2-7} 
 &  5-way & 10-way & \multicolumn{1}{l}{GFSL} & 5-way & 10-way & \multicolumn{1}{l}{GFSL} \\ \hline
\textbf{MGIMN(ours)} & \textbf{87.84} & \textbf{80.60} & \textbf{57.66} & \textbf{54.98} & \textbf{40.12} & \textbf{19.61} \\ \hline
w/o episode & 86.22 & 78.99 & 56.67 & 54.14 & 39.53 & 18.69 \\
w/o class & 84.56 & 76.89 & 54.62 & 54.09 & 39.10 & 17.53 \\
w/o instance & 87.74 & 79.93 & 57.39 & 53.65 & 38.86 & 18.67 \\
w/o instance\&class\&episode &  80.53 & 70.94 & 42.54 & 51.81 & 37.10 & 16.48 \\ \hline
\end{tabular}
\caption{Ablation study results on Liu and Huffpost datasets.} 
\label{ablation}
\end{table*}


\subsection{\textbf{Ablation Study}}
To further validate the effect of different interaction levels and instance matching vector, we make some ablation studies on both the datasets Liu and Huffpost. The settings are totally same with the main experiments.

\subsubsection{Different Interaction Levels}
We respectively take out the single-level interaction layer and see how the specific alignment feature affects the performance. As shown in the Table \ref{ablation}, when taking out the specific interaction layer, the performance decreases in varying degrees, which explains that each alignment layer has positive effect on the performance and can complement each other. It is noted that class-level interaction layer has the greatest impact. The model can pay attention to the whole class context through class-level interaction, which makes the model encode more precise class semantic information. It is the key to judge the relationship between query and class.


\subsubsection{Instance-wise Matching Vector}
We remove all interaction layers in our model, named `w/o instance \& class \& episode' in Table \ref{ablation}. Then it is the same as matching network except that the scalar matching score is replaced by instance-wise matching. 
We make comparison with matching network.
As given in the Table \ref{ablation}, \ref{main_result_1} and \ref{main_result_3}, it performs better than matching network, e.g. Liu dataset improves $3.49\%$ of 10-way 5-shot FSL score and Huffpost dataset improves $2.30\%$ of GFSL score. Unlike scalar comparison in matching network, our approach can make fine-grained instance vector comparisons in fine-grained feature dimensional level.

\subsection{Number of Classes and Inference Speed}
\label{rtc}
As shown in Table \ref{result_rtc}, the inference speed increases linearly with the increase of the number of classes, from $315ms/query$ to $1630ms/query$ when $c$ increases from $50$ to $318$. 
It is challenging for deploying the model to the online application. To address the problem of inference speed, motivated by the idea of retrieval in traditional search system, we propose the retrieval-then-classify method(RTC for short). 
(1)Stage1-retrieval: We construct the class-wise representation by averaging the vector of each support instance, produced by MGIMN encoder and then calculate the similarity between query and class-wise representation. 
In our experiments, we retrieve top $N=10$ classes with $K=5$ instances per class. (2)Stage2-classify: Retrieved support instances by stage1 are classified by MGIMN proposed in this paper. The $C$-way $5$-shot task is reduced to a $10$-way $5$-shot which can greatly save the inference time and computation cost.

\begin{table*}[]
\centering
\begin{tabular}{lcccccc}
\hline
\multirow{2}{*}{Methods} & \multicolumn{2}{c}{Liu(c=50)} & \multicolumn{2}{c}{OOS(c=150)} & \multicolumn{2}{c}{Amzn(c=318)} \\ \cline{2-7} 
 & score & speed & score & speed & score & speed \\ \hline
MGIMN-overall & 57.66 & 315 & 76.23 & 757 & 49.46 & 1630 \\ \hline
RTC-BM25 & 54.97 & 55 & 74.80 & 56 & 44.76 & 58 \\
RTC-oribert & 52.93 & 60 & 70.55 & 65 & 31.09 & 70 \\
RTC-mgimnbert & 56.21 & 60 & 75.58 & 65 & 46.80 & 70 \\ \hline
\end{tabular}
\caption{The generalized FSL accuracy(\%) and inference speed. Speed is reported by averaging for processing $100$ queries and the value is the processing time per query($ms/query$)} 
\label{result_rtc}
\end{table*}

In addition to the generalized FSL metric score, we also report the inference speed (processing time per query) to show the effectiveness of retrieval-then-classify. 
We can see that the inference speed of retrieval-then-classify is greatly increased by $5\times{}$ to $23\times{}$, with a small amount of performance loss. 
At the same time, comparing with other retrieval methods (e.g. BM25 and original bert), our approach can further improve the performance

\section{Related Work}

\subsection{Few-shot Learning}

Intuitively, the few-shot learning focus on learn a classifier only using a few labeled training examples, which is similar to human intelligence. Since the aim of few-shot learning is highly attractive, researchers have proposed various few-shot learning approaches.

Generally, the matching network encodes query text and all support instances independently \cite{vinyals2016matching}, then computes the cosine similarity between query vector and each support sample vector, and finally computes average of all scores for per class. The prototypical network basically encodes query text and support instances independently \cite{snell2017prototypical}, then computes class-wise vector by averaging vectors of all support instances in each class, and finally computes euclidean distances as final scores. \citet{sung2018learning} introduced relation network \cite{sung2018learning} that
exploits a relation module to model relationship between query vector and class-wise vectors rather than leveraging the distance metric (e.g., euclidean and cosine). \citet{geng2019induction} introduced the induction network and leverage the induction module that take the dynamic routing as a vital algorithm
to induce and generalize class-wise representations. 
In the few-shot scenario, the model-agnostic manner usually viewed as the improved version of few-shot, and defined as MAML \cite{DBLP:conf/icml/FinnAL17},
which can be exploited in different fields MT \citep{DBLP:conf/emnlp/GuWCLC18,DBLP:conf/aaai/LiWY20}, dialog generation \citep{DBLP:conf/acl/QianY19,DBLP:conf/emnlp/HuangFMDW20}.

For few-shot text classification, researchers have also proposed various techniques to improve the existing approaches for few-shot learning.
Basically, the one of our strong baselines \textbf{Proto-HATT}  is introduced by \citet{gao2019hybrid}, that leverages the attention with instance-level and feature-level then  highlight both the vital features and support point. \citet{ye2019multi} also tries to encode both query and per support set by leveraging the interactive way at word level with taking the  matching information into account.

\subsection{Text Matching}
Text matching model aims to predict the score of text pair dependent on massive labeled data. Before BERT, related work focuses on deriving the matching information between two text sequences based on the matching aggregation framework. It performs matching in low-level and aggregates matching results based on attention mechanism. 
Many studies are proposed to improve performance. The attend-compare-aggregate method \cite{parikh2016decomposable} which has an effectiveness on alignment, meanwhile aggregates the aligned feature by using feed-forward architecture. 
The previous work extracts fine-grained matching feature with bilateral  matching operation by considering the multi-perspective case\cite{wang2017bilateral}. 
\citet{tay2017compare} exploit the factorization layers to enhance the word representation via scalar features with an effective and strong compressed vectors for alignment.  
\citet{yang2019simple} present a straightforward but efficient text matching model using strong alignment features. 
After the PTM (e.g., BERT) is presented \cite{bert}, it has became commonly used approach on the various fields of NLP. Thus, many text matching methods are also leveraging the PTM. For example, \citet{reimers-gurevych-2019-sentence} use the sentence embeddings of BERT to conduct text matching, and \citet{DBLP:journals/corr/abs-2104-08821} use contrastive learning to train text matching models. 
Additionally, to handle the issue of few-shot learning architecture,
we employ the similar idea of comparison, aggregation and introduce new architecture  multi-grained interactive matching network.

\section{Conclusion}
In this research investigation, we introduce the \textbf{M}ulti-\textbf{g}rained \textbf{I}nteractive \textbf{M}atching \textbf{N}etwork(\textbf{MGIMN}) for the text classification task with few-shot  manner.
Meanwhile, we introduce a two-stage method retrieval-then-classify (RTC) to solve the inference performance in realistic scenery. 
Experiment results illustrate that
the presented method obtains the best result in all five different kinds of datasets with two evaluation metrics. 
Moreover, RTC method obviously make the inference speed getting faster.
We will make further investigations on the the task of domain adaptation problem by extending our proposed method.



\bibliography{main}

\begin{thebibliography}{39}
\expandafter\ifx\csname natexlab\endcsname\relax\def\natexlab#1{#1}\fi

\bibitem[{Bansal et~al.(2020)Bansal, Jha, Munkhdalai, and
  McCallum}]{bansal2020self}
Trapit Bansal, Rishikesh Jha, Tsendsuren Munkhdalai, and Andrew McCallum. 2020.
\newblock Self-supervised meta-learning for few-shot natural language
  classification tasks.
\newblock \emph{arXiv preprint arXiv:2009.08445}.

\bibitem[{Bao et~al.(2019)Bao, Wu, Chang, and Barzilay}]{bao2019few}
Yujia Bao, Menghua Wu, Shiyu Chang, and Regina Barzilay. 2019.
\newblock Few-shot text classification with distributional signatures.
\newblock \emph{arXiv preprint arXiv:1908.06039}.

\bibitem[{Bragg et~al.(2021)Bragg, Cohan, Lo, and Beltagy}]{bragg2021flex}
Jonathan Bragg, Arman Cohan, Kyle Lo, and Iz~Beltagy. 2021.
\newblock Flex: Unifying evaluation for few-shot nlp.
\newblock \emph{arXiv preprint arXiv:2107.07170}.

\bibitem[{Brown et~al.(2020)Brown, Mann, Ryder, Subbiah, Kaplan, Dhariwal,
  Neelakantan, Shyam, Sastry, Askell et~al.}]{brown2020language}
Tom~B Brown, Benjamin Mann, Nick Ryder, Melanie Subbiah, Jared Kaplan, Prafulla
  Dhariwal, Arvind Neelakantan, Pranav Shyam, Girish Sastry, Amanda Askell,
  et~al. 2020.
\newblock Language models are few-shot learners.
\newblock \emph{arXiv preprint arXiv:2005.14165}.

\bibitem[{Devlin et~al.(2019)Devlin, Chang, Lee, and Toutanova}]{bert}
Jacob Devlin, Ming{-}Wei Chang, Kenton Lee, and Kristina Toutanova. 2019.
\newblock {BERT:} pre-training of deep bidirectional transformers for language
  understanding.
\newblock In \emph{Proceedings of the 2019 Conference of the North American
  Chapter of the Association for Computational Linguistics: Human Language
  Technologies}, pages 4171--4186.

\bibitem[{Finn et~al.(2017)Finn, Abbeel, and Levine}]{DBLP:conf/icml/FinnAL17}
Chelsea Finn, Pieter Abbeel, and Sergey Levine. 2017.
\newblock Model-agnostic meta-learning for fast adaptation of deep networks.
\newblock In \emph{Proceedings of the 34th International Conference on Machine
  Learning}, pages 1126--1135.

\bibitem[{Gao et~al.(2020)Gao, Fisch, and Chen}]{gao2020making}
Tianyu Gao, Adam Fisch, and Danqi Chen. 2020.
\newblock Making pre-trained language models better few-shot learners.
\newblock \emph{arXiv preprint arXiv:2012.15723}.

\bibitem[{Gao et~al.(2019)Gao, Han, Liu, and Sun}]{gao2019hybrid}
Tianyu Gao, Xu~Han, Zhiyuan Liu, and Maosong Sun. 2019.
\newblock Hybrid attention-based prototypical networks for noisy few-shot
  relation classification.
\newblock In \emph{Proceedings of the AAAI Conference on Artificial
  Intelligence}, volume~33, pages 6407--6414.

\bibitem[{Gao et~al.(2021)Gao, Yao, and
  Chen}]{DBLP:journals/corr/abs-2104-08821}
Tianyu Gao, Xingcheng Yao, and Danqi Chen. 2021.
\newblock Simcse: Simple contrastive learning of sentence embeddings.
\newblock \emph{arXiv preprint arXiv:2104.08821}.

\bibitem[{Geng et~al.(2019)Geng, Li, Li, Zhu, Jian, and
  Sun}]{geng2019induction}
Ruiying Geng, Binhua Li, Yongbin Li, Xiaodan Zhu, Ping Jian, and Jian Sun.
  2019.
\newblock Induction networks for few-shot text classification.
\newblock \emph{arXiv preprint arXiv:1902.10482}.

\bibitem[{Gu et~al.(2018)Gu, Wang, Chen, Li, and
  Cho}]{DBLP:conf/emnlp/GuWCLC18}
Jiatao Gu, Yong Wang, Yun Chen, Victor O.~K. Li, and Kyunghyun Cho. 2018.
\newblock Meta-learning for low-resource neural machine translation.
\newblock In \emph{Proceedings of the 2018 Conference on Empirical Methods in
  Natural Language Processing}, pages 3622--3631.

\bibitem[{Han et~al.(2018)Han, Zhu, Yu, Wang, Yao, Liu, and
  Sun}]{han2018fewrel}
Xu~Han, Hao Zhu, Pengfei Yu, Ziyun Wang, Yuan Yao, Zhiyuan Liu, and Maosong
  Sun. 2018.
\newblock Fewrel: A large-scale supervised few-shot relation classification
  dataset with state-of-the-art evaluation.
\newblock \emph{arXiv preprint arXiv:1810.10147}.

\bibitem[{Huang et~al.(2020)Huang, Feng, Ma, Du, and
  Wu}]{DBLP:conf/emnlp/HuangFMDW20}
Yi~Huang, Junlan Feng, Shuo Ma, Xiaoyu Du, and Xiaoting Wu. 2020.
\newblock Towards low-resource semi-supervised dialogue generation with
  meta-learning.
\newblock In \emph{Findings of the Association for Computational Linguistics:
  {EMNLP} 2020}, pages 4123--4128.

\bibitem[{Karan and {\v{S}}najder(2016)}]{karan2016faqir}
Mladen Karan and Jan {\v{S}}najder. 2016.
\newblock Faqir--a frequently asked questions retrieval test collection.
\newblock In \emph{International Conference on Text, Speech, and Dialogue},
  pages 74--81. Springer.

\bibitem[{Kim(2014)}]{DBLP:conf/emnlp/Kim14}
Yoon Kim. 2014.
\newblock Convolutional neural networks for sentence classification.
\newblock In \emph{Proceedings of the 2014 Conference on Empirical Methods in
  Natural Language Processing}, pages 1746--1751.

\bibitem[{Krone et~al.(2020)Krone, Zhang, and Diab}]{krone2020learning}
Jason Krone, Yi~Zhang, and Mona Diab. 2020.
\newblock Learning to classify intents and slot labels given a handful of
  examples.
\newblock \emph{arXiv preprint arXiv:2004.10793}.

\bibitem[{Larson et~al.(2019)Larson, Mahendran, Peper, Clarke, Lee, Hill,
  Kummerfeld, Leach, Laurenzano, Tang, and Mars}]{larson-etal-2019-evaluation}
Stefan Larson, Anish Mahendran, Joseph~J. Peper, Christopher Clarke, Andrew
  Lee, Parker Hill, Jonathan~K. Kummerfeld, Kevin Leach, Michael~A. Laurenzano,
  Lingjia Tang, and Jason Mars. 2019.
\newblock \href {https://doi.org/10.18653/v1/D19-1131} {An evaluation dataset
  for intent classification and out-of-scope prediction}.
\newblock In \emph{Proceedings of the 2019 Conference on Empirical Methods in
  Natural Language Processing and the 9th International Joint Conference on
  Natural Language Processing (EMNLP-IJCNLP)}, pages 1311--1316, Hong Kong,
  China. Association for Computational Linguistics.

\bibitem[{Li et~al.(2020{\natexlab{a}})Li, Huang, Lan, Feng, Li, and
  Wang}]{li2020boosting}
Aoxue Li, Weiran Huang, Xu~Lan, Jiashi Feng, Zhenguo Li, and Liwei Wang.
  2020{\natexlab{a}}.
\newblock Boosting few-shot learning with adaptive margin loss.
\newblock In \emph{Proceedings of the IEEE/CVF Conference on Computer Vision
  and Pattern Recognition}, pages 12576--12584.

\bibitem[{Li et~al.(2020{\natexlab{b}})Li, Wang, and
  Yu}]{DBLP:conf/aaai/LiWY20}
Rumeng Li, Xun Wang, and Hong Yu. 2020{\natexlab{b}}.
\newblock Metamt, a meta learning method leveraging multiple domain data for
  low resource machine translation.
\newblock In \emph{The Thirty-Fourth {AAAI} Conference on Artificial
  Intelligence}.

\bibitem[{Liu et~al.(2021)Liu, Yuan, Fu, Jiang, Hayashi, and
  Neubig}]{liu2021pre}
Pengfei Liu, Weizhe Yuan, Jinlan Fu, Zhengbao Jiang, Hiroaki Hayashi, and
  Graham Neubig. 2021.
\newblock Pre-train, prompt, and predict: A systematic survey of prompting
  methods in natural language processing.
\newblock \emph{arXiv preprint arXiv:2107.13586}.

\bibitem[{Liu et~al.(2019)Liu, Eshghi, Swietojanski, and
  Rieser}]{liu2019benchmarking}
Xingkun Liu, Arash Eshghi, Pawel Swietojanski, and Verena Rieser. 2019.
\newblock Benchmarking natural language understanding services for building
  conversational agents.
\newblock \emph{arXiv preprint arXiv:1903.05566}.

\bibitem[{Luo et~al.(2021)Luo, Liu, Lin, and Zhang}]{luo2021don}
Qiaoyang Luo, Lingqiao Liu, Yuhao Lin, and Wei Zhang. 2021.
\newblock Don’t miss the labels: Label-semantic augmented meta-learner for
  few-shot text classification.
\newblock In \emph{Findings of the Association for Computational Linguistics:
  ACL-IJCNLP 2021}, pages 2773--2782.

\bibitem[{Misra.(2018)}]{huffpost}
Rishabh Misra. 2018.
\newblock Huffpost news category dataset.

\bibitem[{Nguyen et~al.(2020)Nguyen, Zhang, Xia, and Yu}]{nguyen2020dynamic}
Hoang Nguyen, Chenwei Zhang, Congying Xia, and Philip~S Yu. 2020.
\newblock Dynamic semantic matching and aggregation network for few-shot intent
  detection.
\newblock \emph{arXiv preprint arXiv:2010.02481}.

\bibitem[{Parikh et~al.(2016)Parikh, T{\"a}ckstr{\"o}m, Das, and
  Uszkoreit}]{parikh2016decomposable}
Ankur~P Parikh, Oscar T{\"a}ckstr{\"o}m, Dipanjan Das, and Jakob Uszkoreit.
  2016.
\newblock A decomposable attention model for natural language inference.
\newblock \emph{arXiv preprint arXiv:1606.01933}.

\bibitem[{Qian and Yu(2019)}]{DBLP:conf/acl/QianY19}
Kun Qian and Zhou Yu. 2019.
\newblock Domain adaptive dialog generation via meta learning.
\newblock In \emph{Proceedings of the 57th Conference of the Association for
  Computational Linguistics}, pages 2639--2649.

\bibitem[{Reimers and Gurevych(2019)}]{reimers-gurevych-2019-sentence}
Nils Reimers and Iryna Gurevych. 2019.
\newblock Sentence-{BERT}: Sentence embeddings using {S}iamese {BERT}-networks.
\newblock In \emph{Proceedings of the 2019 Conference on Empirical Methods in
  Natural Language Processing}, pages 3982--3992.

\bibitem[{Schick and Sch{\"u}tze(2020)}]{schick2020s}
Timo Schick and Hinrich Sch{\"u}tze. 2020.
\newblock It's not just size that matters: Small language models are also
  few-shot learners.
\newblock \emph{arXiv preprint arXiv:2009.07118}.

\bibitem[{Snell et~al.(2017)Snell, Swersky, and Zemel}]{snell2017prototypical}
Jake Snell, Kevin Swersky, and Richard~S Zemel. 2017.
\newblock Prototypical networks for few-shot learning.
\newblock \emph{arXiv preprint arXiv:1703.05175}.

\bibitem[{Sung et~al.(2018)Sung, Yang, Zhang, Xiang, Torr, and
  Hospedales}]{sung2018learning}
Flood Sung, Yongxin Yang, Li~Zhang, Tao Xiang, Philip~HS Torr, and Timothy~M
  Hospedales. 2018.
\newblock Learning to compare: Relation network for few-shot learning.
\newblock In \emph{Proceedings of the IEEE conference on computer vision and
  pattern recognition}, pages 1199--1208.

\bibitem[{Tay et~al.(2017)Tay, Tuan, and Hui}]{tay2017compare}
Yi~Tay, Luu~Anh Tuan, and Siu~Cheung Hui. 2017.
\newblock Compare, compress and propagate: Enhancing neural architectures with
  alignment factorization for natural language inference.
\newblock \emph{arXiv preprint arXiv:1801.00102}.

\bibitem[{Vaswani et~al.(2017)Vaswani, Shazeer, Parmar, Uszkoreit, Jones,
  Gomez, Kaiser, and Polosukhin}]{VaswaniEtAl17-Transformer}
Ashish Vaswani, Noam Shazeer, Niki Parmar, Jakob Uszkoreit, Llion Jones,
  Aidan~N. Gomez, Lukasz Kaiser, and Illia Polosukhin. 2017.
\newblock Attention is all you need.
\newblock In \emph{Advances in Neural Information Processing Systems}, pages
  5998--6008.

\bibitem[{Vinyals et~al.(2016)Vinyals, Blundell, Lillicrap, Kavukcuoglu, and
  Wierstra}]{vinyals2016matching}
Oriol Vinyals, Charles Blundell, Timothy Lillicrap, Koray Kavukcuoglu, and Daan
  Wierstra. 2016.
\newblock Matching networks for one shot learning.
\newblock \emph{arXiv preprint arXiv:1606.04080}.

\bibitem[{Wang et~al.(2017)Wang, Hamza, and Florian}]{wang2017bilateral}
Zhiguo Wang, Wael Hamza, and Radu Florian. 2017.
\newblock Bilateral multi-perspective matching for natural language sentences.
\newblock \emph{arXiv preprint arXiv:1702.03814}.

\bibitem[{Yang et~al.(2019)Yang, Zhang, Gao, Ji, and Chen}]{yang2019simple}
Runqi Yang, Jianhai Zhang, Xing Gao, Feng Ji, and Haiqing Chen. 2019.
\newblock Simple and effective text matching with richer alignment features.
\newblock \emph{arXiv preprint arXiv:1908.00300}.

\bibitem[{Ye and Ling(2019)}]{ye2019multi}
Zhi-Xiu Ye and Zhen-Hua Ling. 2019.
\newblock Multi-level matching and aggregation network for few-shot relation
  classification.
\newblock \emph{arXiv preprint arXiv:1906.06678}.

\bibitem[{Yin(2020)}]{yin2020meta}
Wenpeng Yin. 2020.
\newblock Meta-learning for few-shot natural language processing: A survey.
\newblock \emph{arXiv preprint arXiv:2007.09604}.

\bibitem[{Yury.(2020)}]{amzn}
Kashnitsky Yury. 2020.
\newblock Hierarchical text classification of amazon product reviews.

\bibitem[{Zhou et~al.(2016)Zhou, Qi, Zheng, Xu, Bao, and
  Xu}]{zhou-etal-2016-text}
Peng Zhou, Zhenyu Qi, Suncong Zheng, Jiaming Xu, Hongyun Bao, and Bo~Xu. 2016.
\newblock Text classification improved by integrating bidirectional {LSTM} with
  two-dimensional max pooling.
\newblock In \emph{Proceedings of {COLING} 2016, the 26th International
  Conference on Computational Linguistics: Technical Papers}, pages 3485--3495.

\end{thebibliography}
\bibliographystyle{acl_natbib}

\end{CJK}
\end{document}